\documentclass[10pt,letterpaper]{article}
\usepackage[top=0.85in,left=2.75in,footskip=0.75in]{geometry}

\usepackage{amsmath,amssymb}

\usepackage{changepage}

\usepackage{textcomp,marvosym}

\usepackage{cite}

\usepackage{nameref,hyperref}

\usepackage[right]{lineno}

\usepackage[nopatch=eqnum]{microtype}
\DisableLigatures[f]{encoding = *, family = * }

\usepackage[table]{xcolor}

\usepackage{array}

\usepackage{booktabs}
\usepackage[ruled]{algorithm2e}

\newcolumntype{+}{!{\vrule width 2pt}}

\newlength\savedwidth

\raggedright
\usepackage[aboveskip=1pt,labelfont=bf,labelsep=period,justification=raggedright,singlelinecheck=off]{caption}

\makeatletter
\renewcommand{\@biblabel}[1]{\quad#1.}
\makeatother

\usepackage{lastpage,fancyhdr,graphicx}
\usepackage{epstopdf}
\fancyheadoffset[L]{2.25in}
\fancyfootoffset[L]{2.25in}
\begin{document}
\vspace*{0.2in}

\begin{flushleft}
{\Large
\textbf\newline{DiffDesign: Controllable Diffusion with Meta Prior for Efficient Interior Design Generation} 
}
\newline
\\
Tao Geng\textsuperscript{1,*},
Yuxuan Yang\textsuperscript{1} 
\\
\bigskip
\textbf{1} Nanjing Forestry University, Nanjing, China
\\
\bigskip






* Correspondence: nfugt@njfu.edu.cn

\end{flushleft}
\section*{Abstract}
Interior design is a complex and creative discipline involving aesthetics, functionality, ergonomics, and materials science. Effective solutions must meet diverse requirements, typically producing multiple deliverables such as renderings and design drawings from various perspectives. Consequently, interior design processes are often inefficient and demand significant creativity. With advances in machine learning, generative models have emerged as a promising means of improving efficiency by creating designs from text descriptions or sketches. However, few generative works focus on interior design, leading to substantial discrepancies between outputs and practical needs, such as differences in size, spatial scope, and the lack of controllable generation quality. To address these challenges, we propose DiffDesign, a controllable diffusion model with meta priors for efficient interior design generation. Specifically, we utilize the generative priors of a 2D diffusion model pre-trained on a large image dataset as our rendering backbone. We further guide the denoising process by disentangling cross-attention control over design attributes, such as appearance, pose, and size, and introduce an optimal transfer-based alignment module to enforce view consistency. Simultaneously, we construct an interior design-specific dataset, DesignHelper, consisting of over 400 solutions across more than 15 spatial types and 15 design styles. This dataset helps fine-tune DiffDesign. Extensive experiments conducted on various benchmark datasets demonstrate the effectiveness and robustness of DiffDesign.



\section{Introduction}
\label{sec:1}
With globalization and urbanization, the demand for comfortable, functional, and aesthetically pleasing living and working environments has surged, driving exponential growth in interior design (ID) services~\cite{bittencourt2015usability}. As a complex and creative discipline~\cite{huang2020category,kalantari2020virtual}, ID encompasses aesthetics, functionality, and materials science to address diverse needs. Despite notable advancements~\cite{sydor2021mycelium,wang2024comprehensive,rashdan2016impact,wang2023awesome}, ID still faces challenges~\cite{kalantari2020virtual}, such as the mismatch between growing demand and existing design methods, the complexity of the design process, inefficiencies from frequent changes, and designers' reliance on conventional methods, limiting innovation. Additionally, designers must integrate interdisciplinary knowledge, leverage new technological tools, consider sustainability, and adapt to varied cultural and regional needs~\cite{wilson2015design}. To enhance efficiency and foster innovation, the industry must explore new design methods, develop designers' skills comprehensively, and effectively utilize modern technology.

The advent of machine learning has introduced generative models as effective tools for enhancing design efficiency~\cite{liu2023sounding,wang2024image,koksal2023controllable,wang2024towards,buildings14061528}. Gruver et al.\cite{gruver2024protein} used diffusion models to design protein structures, while Cao et al.\cite{cao2023difffashion} employed diffusion to combine reference appearance images and clothing images for fashion design. Generative models, including generative adversarial networks (GANs)\cite{creswell2018generative,goodfellow2020generative}, variational autoencoders (VAEs)\cite{doersch2016tutorial,sonderby2016ladder}, and diffusion models~\cite{croitoru2023diffusion,yang2023diffusion, Chen2023, Chen2024,Chen20242}, can autonomously generate novel design solutions by learning from extensive design data, thereby reducing designers' workload and offering increased inspiration and creative possibilities~\cite{li2018generative}.

\begin{figure*}[!h]
    \centering
    \includegraphics[width=\textwidth]{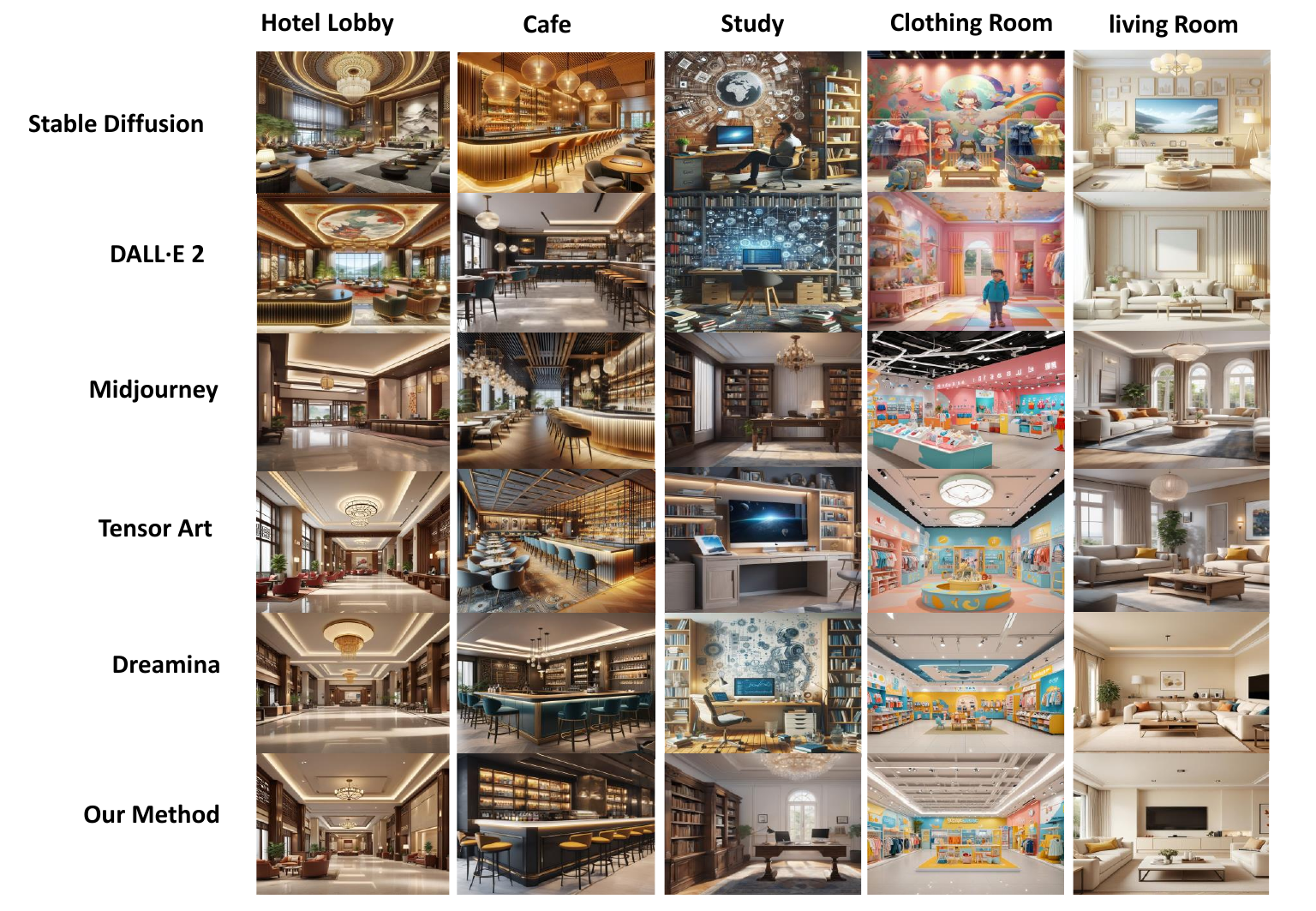}
    \caption{\textbf{Interior design images generated by different models}. Stable Diffusion has issues with incorrect ceiling perspective, haphazard furniture placement, and a cluttered distant scene lacking proper shading. DALL·E 2 produces duplicated, unrealistic objects, with distorted furniture perspectives and cluttered display cases. Midjourney suffers from blurred ceiling patterns, disproportionate lamp sizes, and illogical window placements merged with railing patterns. Tensor Art features unreasonable spatial dimensions, low ceilings, oversized furniture, and overly intense lighting that lacks realism. \textbf{Dreamina} displays inexplicable decorations, blended, stuck-together, and cartoonish items. None of these images are up to the interior design requirements but our method does.}
    \label{fig:intro}
\end{figure*}

However, the practical application of interior design still faces significant challenges~\cite{sydora2020rule}. First, the renderings in this field must accurately reflect every detail of the drawings, including furniture dimensions, electrical layouts, and pipeline directions, to prevent errors and rework during construction. Existing methods, however, often produce uncontrollable results that fail to meet the required level of professionalism and accuracy~\cite{maclean2020questions,mitra2016fundamentals}. Second, interior design renderings typically include floor plans, elevations, sections, and 3D views to convey spatial design from multiple perspectives. Yet, few works specifically focus on generating these types of drawings and renderings, resulting in a substantial gap between generated outputs and actual needs~\cite{zhang2023generative,hu2023exploring,wang2023hacking}. For instance, the generated outputs may exhibit discrepancies in size, spatial range, or lack essential layout elements such as furniture and decorations. Fig~\ref{fig:intro} provides examples of image generation by existing methods, illustrating these shortcomings. Additionally, interior design plans often require a degree of flexibility to accommodate changes in client needs and construction adjustments~\cite{estaji2017review}. This necessitates diverse generated outputs that can meet various complex requirements. These challenges make image and model generation for interior design particularly difficult.

To address these challenges, we propose DiffDesign, a controllable diffusion model with a meta prior for efficient interior design generation. DiffDesign is a novel text-to-image approach that enhances 2D diffusion models to synthesize high-quality interior design drawings from minimal prior descriptions, such as a single word or sketch. Specifically, we utilize a pre-trained 2D diffusion model as our rendering backbone and incorporate a cross-view attention module, as used in video diffusion models, to ensure regional consistency between input text and generated images. Furthermore, to generate controllable and reliable renderings, we decompose the task into explicitly disentangled control of appearance and design specifications. For appearance control, we derive semantic context from trainable reference examples, locking the feature extractor while activating the attention module to provide layer-by-layer contextual guidance. This approach retains the capabilities of the large diffusion model while generating images with preserved reference features and textual descriptions. For design specifications, we implement image control by adding design text description attention to the locked UNet decoder. Finally, we integrate these two aspects of control into the diffusion model as a meta prior to generate high-quality, controllable interior design renderings.

Additionally, we construct a specialized dataset, called DesignHelper, to support the training and evaluation of DiffDesign. Specifically, we selected over 15 space types and 15 design styles, collecting more than 600 design solution pairs. Each solution includes a textual description provided by five design-focused researchers, along with floor plans and renderings. Extensive experiments, including both machine verification and human evaluation, demonstrate the reliability of our approach, i.e., the images generated by DiffDesign meet industry standards and accurately depict the design requirements specified in the prompts.

In summary, our contributions are mainly as follows:
\begin{itemize}
    \item We explore and address three challenges of generative models in interior design, providing ideas for the automatic and controllable generation of interior design solutions.
    \item We propose DiffDesign, a controllable Diffusion with meta prior for interior design generation. It decouples the control of appearance and ID specifications and uses them as a meta-prior to guide the generation.
    \item We conduct an ID-specialized dataset, DesignHelper, to support the training and evaluation of DiffDesign. It comprises over 15 space types and 15 design styles, resulting in more than 400 design solution pairs.
    \item Extensive experiments demonstrate that DiffDesign outperforms various previous methods both quantitatively and qualitatively in terms of visual quality, diversity, and professionalism in interior design.
\end{itemize}

\section{Related work}
\label{sec:2}
In this section, we briefly summarize the foundations of interior design and Diffusion models.

\subsection{Interior Design}
\label{sec:2.1}
Interior design \cite{bittencourt2015usability,huang2020category} refers to the process which designers use their knowledge of art and engineering to create interior spaces that reflect specific decorative styles for clients. Designers must choose appropriate design elements (e.g., suitable tiles, furniture, colors, and patterns) to shape these styles. A strong decorative style is essential for making the design unique \cite{kalantari2020virtual,sydor2021mycelium}. Traditionally, designers first construct design drawings based on client needs, then create two-dimensional (2D) drawings and corresponding three-dimensional (3D) models. Next, materials are mapped onto the 3D model, lighting is arranged, and renderings are generated. Ultimately, clients evaluate the design's suitability based on these renderings.

However, this linear, traditional workflow in interior design is inefficient, requiring designers to spend significant time drawing designs without real-time communication with clients, leading to multiple revisions \cite{rashdan2016impact,wilson2015design}. Additionally, if clients are dissatisfied at any stage, designers must redo the entire design. Furthermore, the traditional workflow is cumbersome and stifles creative design. On one hand, in pursuit of efficiency, designers may adopt rigid methods to quickly generate ideas. On the other hand, even if designers have abundant creative inspiration, translating these ideas into renderings requires considerable manpower, and they can only realize a portion of their concepts within limited time constraints. Therefore, facilitating the rapid generation of diverse interior design renderings is essential for overcoming inefficiencies and promoting creativity in the design process. To tackle this issue, this study aims to enhance design efficiency and address creativity challenges by developing a diffusion model specifically for interior design generation. Additionally, this study constructs a specialized dataset for interior design solutions to support model training and scheme generation.

\subsection{Diffusion Models}
\label{sec:2.2}
Diffusion models (DMs) learn to reverse the diffusion process by progressively denoising samples from a normal distribution~\cite{croitoru2023diffusion,kingma2021variational}. This approach enables diffusion-based generative models to produce high-fidelity images from various text prompts~\cite{cao2023masactrl,dhariwal2021diffusion}. However, DMs operating in pixel space suffer from high generation latency, significantly limiting their practical applications~\cite{xiao2021tackling}. The Latent Diffusion Model (LDM)~\cite{rombach2022high} addressed this issue by training a Variational Auto-Encoder (VAE) to encode pixel space into a latent space, where the DM is then applied. This method reduces computational costs significantly while maintaining image quality, thus enhancing the usability of DMs. Following this, the latest and most advanced open-source version is SD-XL~\cite{podell2023sdxl}, which significantly outperforms its predecessors. In this work, we use SD-XL as our default backbone.

Recently, researchers have invested significant effort in enhancing the performance of DMs \cite{hong2022headnerf,li2024blip}. The current DMs can be categorized into two types: (i) Efficient sampling techniques \cite{lin20223d} which aim to reduce the number of required denoising steps. Related methods \cite{liu2023labelled,liu2023conditional} has successfully minimized the denoising steps to just one through iterative distillation, where the number of steps is halved iteratively; and (ii) Architectural compression \cite{yang2021joint,deng2019accurate} which focuses on making each sampling step more efficient. Recent methods \cite{ho2022classifier,or2022stylesdf} achieve this by eliminating multiple ResNet and attention blocks in the U-Net via distillation, effectively lower computational costs and maintain good image quality. However, they require retraining the DM backbone to enhance efficiency, necessitating thousands of A100 GPU hours \cite{poole2022dreamfusion}. 
However, although existing methods have achieved good results in image generation, they still face many difficulties in real-life interior design scenarios. These methods ignore the strict regulations and high accuracy requirements in interior design, resulting in a large gap between the generated samples and the actual needs, such as differences in size and spatial range, and the generation quality is uncontrollable.
Meanwhile, recent advances in prompt tuning and controllable diffusion models, such as DreamBooth \cite{ruiz2022dreambooth}, BLIP-2 \cite{li2023blip2}, and GIT \cite{wang2022git} have paved the way for fine-grained control in image generation. However, these models that focus primarily on textual or visual domain adaptation and still face challenges in real-life interior design scenarios.
In this work, we focus on interior design generation and aim to solve the above problems. The proposed DiffDesign framework introduces a dual-control mechanism that simultaneously addresses both appearance-level and specification-level design control for interior design generation.

\section{Preliminaries}
\label{sec:3}

In this section, we first introduce the problem settings and notation of latent Diffusion models. Next, we introduce the optimization process with variational inference.

\subsection{Latent Diffusion Models.}
\label{sec:3.1}
Diffusion models~\cite{hong2022headnerf,liu2023labelled,liu2023conditional} are generative models designed to synthesize desired data samples from Gaussian noise by iteratively removing noise. Latent diffusion models~\cite{rombach2022high} are a class of diffusion models that operate in the encoded latent space of an autoencoder, denoted as $\mathcal{D}(\mathcal{E}(\cdot))$, where $\mathcal{E}(\cdot)$ represents the encoder and $\mathcal{D}(\cdot)$ represents the decoder. Specifically, given an image $ I $ and a text condition $T_{\text{text}} $, the encoded image latent $z_0 = \mathcal{E}(I)$ is diffused over $T$ time steps into a Gaussian-distributed $z_T \sim \mathcal{N}(0, 1)$.

The model is then trained to learn the reverse denoising process with the following objective:
\begin{equation}\label{eq:obj}
    \mathcal{L} = \mathbb{E}_{z_0, c_{\text{text}}, t, \epsilon \sim \mathcal{N}(0,1)} \bigg[ \Big\lVert \epsilon - \epsilon_\theta \big(z_t, c_{\text{text}}, t\big) \Big\lVert_2^2 \bigg],
\end{equation}
where $\epsilon_\theta$ is parameterized as a trainable U-Net architecture, consisting of layers with intervened convolutions (ResBlock) and self-/cross-attentions (TransBlock). In this paper, we focus on generating interior design and develop our network as a plug-and-play module compatible with the latest state-of-the-art and classic text-to-image latent diffusion model. By integrating our module with Diffusion models, we aim to harness its robust latent space manipulation capabilities to improve overall text-to-image synthesis performance.

\subsection{Optimization with Variational Inference}
\label{sec:3.2}
The optimization of diffusion models can be achieved through variational inference. It is an approximation method for inferring complex distributions, commonly used in probabilistic graphical models. In the context of diffusion models, variational inference helps us optimize model parameters by minimizing the Evidence Lower Bound (ELBO).

First, diffusion models are a class of generative models that generate desired data samples from Gaussian noise by gradual denoising. This is achieved through two processes: the forward diffusion process and the reverse denoising process. In the forward diffusion process, noise is gradually added to the data until it eventually approximates a Gaussian distribution. This process is typically defined as:
\begin{equation}\label{eq:forward}
q(z_t | z_{t-1}) = \mathcal{N}(z_t; \sqrt{\alpha_t} z_{t-1}, (1 - \alpha_t)I),
\end{equation}
where $z_0$ is the original data, $z_t$ is the noisy data at step $t$, and $\alpha_t$ is the noise scale at time step $t$.

Next, in the reverse denoising process, the model attempts to generate the original data from the noisy data by denoising. We need to learn a reverse denoising distribution:
\begin{equation}\label{eq:reverse}
p_\theta(z_{t-1} | z_t), 
\end{equation}
where $\theta$ are the model parameters. To optimize this process, we use variational inference to approximate complex distributions by minimizing the variational lower bound.

To approximate the true data distribution by maximizing $p_\theta(x)$, we define the Evidence Lower Bound (ELBO) as the objective function:
\begin{equation}\label{eq:elbo}
\begin{aligned}
\log p_\theta(x) &= \log \int p_\theta(x, z_{0:T}) \, dz_{0:T} = \log \int q(z_{0:T} | x) \frac{p_\theta(x, z_{0:T})}{q(z_{0:T} | x)} \, dz_{0:T}  \\
&\geq \mathbb{E}_q \left[ \log \frac{p_\theta(x, z_{0:T})}{q(z_{1:T} | x)} \right]=\text{ELBO},
\end{aligned}
\end{equation}
where $q(z_{1:T} | x)$ is the joint distribution of the forward process, and $p_\theta(x, z_{0:T})$ is the joint distribution of the reverse process. Through variational inference, we derive an optimization process that aims to minimize the following loss function:
\begin{equation}\label{eq:loss}
\begin{aligned}
    \text{ELBO} &= \mathbb{E}_{q(z_{0:T} | x)} \left[ \log \frac{p_\theta(x, z_{0:T})}{q(z_{0:T} | x)} \right] \\
    &= \mathbb{E}_{q(z_{0:T} | x)} \left[ \log p_\theta(x | z_0) + \sum_{t=1}^T \log p_\theta(z_{t-1} | z_t) - \sum_{t=1}^T \log q(z_t | z_{t-1}) - \log q(z_0 | x) \right] \\
    &= \mathbb{E}_{q(z_{0:T} | x)} \left[ \log p_\theta(x | z_0) \right] - \sum_{t=1}^T \mathbb{E}_{q(z_t | x)} \left[ D_{\text{KL}} \left( q(z_{t-1} | z_t) \| p_\theta(z_{t-1} | z_t) \right) \right]\\
    &= \mathbb{E}_q \left[ \sum_{t=1}^T D_{\text{KL}} \left( q(z_t | z_{t-1}) \| p_\theta(z_t | z_{t-1}) \right) - \log p_\theta(z_0 | z_1) \right],
\end{aligned}
\end{equation}
where $D_{\text{KL}}$ represents the Kullback-Leibler divergence \cite{dhariwal2021diffusion}, measuring the difference between two distributions. Finally, combining the above, we define the optimization objective of the diffusion model as Eq.\ref{eq:obj}. Through these steps, diffusion models optimized by variational inference can efficiently generate high-quality data samples from Gaussian noise. This optimization method not only improves generation efficiency but also ensures high fidelity of the generated data.


\section{Methodology}
\label{sec:4}
To generate high-quality interior design solutions while overcoming the challenges faced by existing models, we introduce DiffDesign, a controllable diffusion framework with a meta prior for efficient interior design generation. The core goal of DiffDesign is to address three key challenges during the design process: (i) ensuring that the generated designs meet the high standards of interior design, (ii) increasing the diversity of the generated solutions while maintaining compliance with design standards, and (iii) enabling the model to learn general prior knowledge from limited data and text-based guidance, allowing for rapid convergence and efficient generation, even in previously unseen scenarios.
In this section, we provide an overview of DiffDesign (Section \ref{sec:4.1}), which includes a description of the learning process and model architecture. We then explain how the framework controls the generated appearance and design specifications, respectively, in Sections \ref{sec:4.2} and \ref{sec:4.3}. Finally, we discuss the optimization process of DiffDesign with meta priors in Section \ref{sec:4.4}.
The overall framework of DiffDesign is illustrated in Figure \ref{fig:1}, and the pseudo-code of ST-F2M is shown in Algorithm \ref{alg:1}.

\begin{algorithm}
\caption{Overview of DiffDesign}
\label{alg:1}
\KwIn{Text description $c_{\text{text}}$, Reference image features $c_{\text{ref}}$, Design text features $c_{\text{design}}$}
\KwOut{Generated interior design image}

\textbf{Stage 1: Text Encoding}

\Begin{
    Collect manually reviewed text descriptions related to interior design. \\
    Train a two-layer MLP on pre-processed data to dynamically identify key regions using a patch-based screening mechanism. \\
    Use pre-trained CLIP text encoder to extract features from $c_{\text{text}}$. \\
    Calculate weight matrix, emphasizing interior design terms.
}

\textbf{Stage 2: Text-to-Image Generation}

\Begin{
    Integrate text features into a latent diffusion model for efficient generation.\\

    \textbf{Module 1: Generative Appearance Control}

    \Begin{
        Derive semantic appearance context from $c_{\text{text}}$. \\
        Lock feature extractor to maintain reference characteristics. \\
        Activate attention module for contextual guidance. \\
        Integrate $c_{\text{text}}$ into UNet backbone for appearance control using attention.
    }

    \textbf{Module 2: Design Specification Control}

    \Begin{
        Encode design text $c_{\text{design}}$. \\
        Fuse text features into UNet decoder at each layer using attention to ensure design control. \\
        Update layer representations to incorporate design specifications.
    }

    \textbf{Module 3: Final Optimization}

    \Begin{
        Optimize diffusion model using $c_{\text{text}}$, $c_{\text{ref}}$, and $c_{\text{design}}$. \\
        Minimize loss by combining text, reference image, and design features for controllable image generation.
    }
}
\end{algorithm}

\subsection{Overview of DiffDesign}
\label{sec:4.1}
\begin{figure*}[h!]
    \centering
    \includegraphics[width=\textwidth]{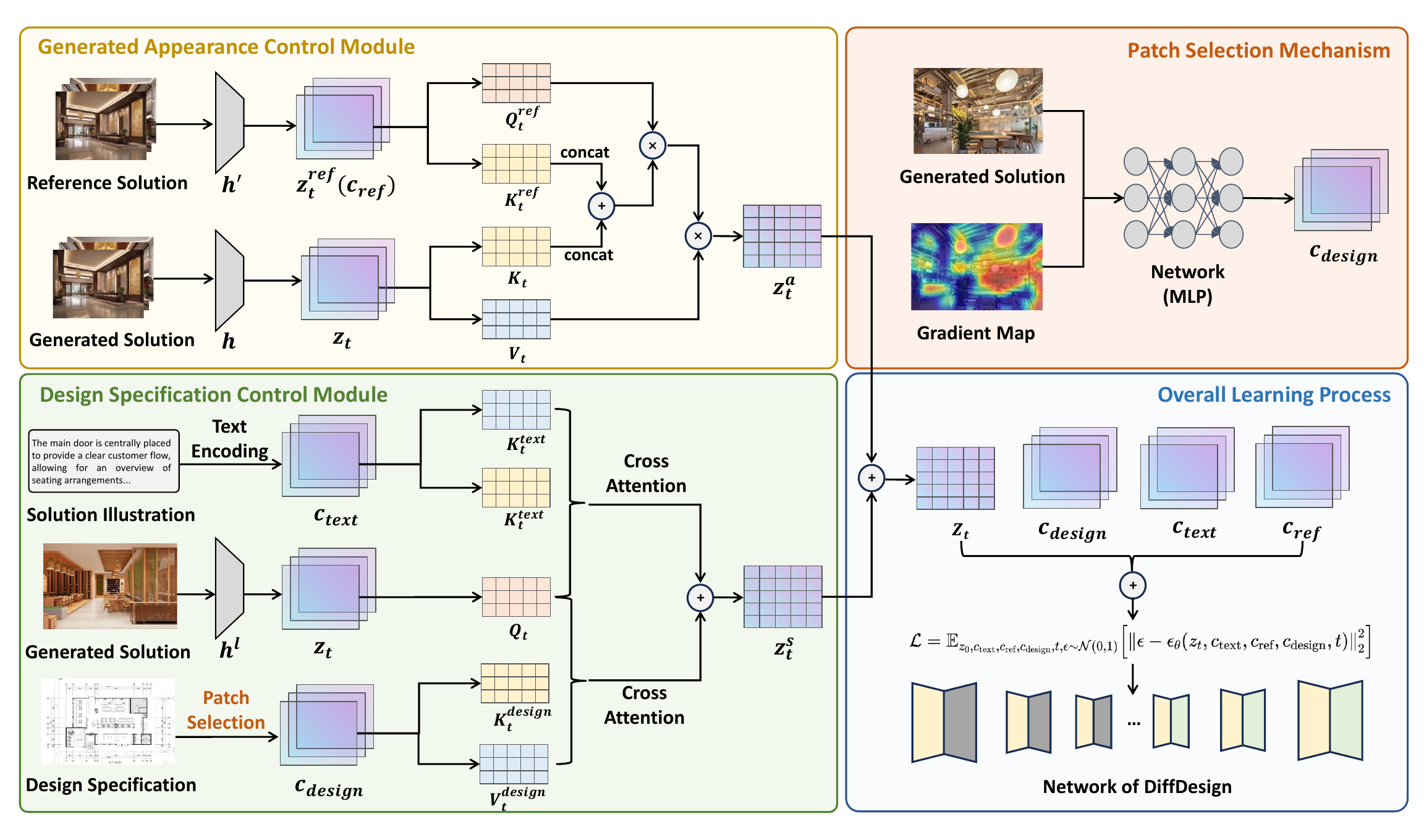}
    \caption{Illustration of the proposed DiffDesign.}
    \label{fig:1}
\end{figure*}

The proposed DiffDesign framework consists of two main stages: the text encoding stage and the text-to-image generation stage. The text-to-image generation stage is further divided into three modules: the appearance control module, the design specification control module, and the final optimization module. This subsection provides a brief overview of the process and the function of each module. Figure \ref{fig:1} provides the framework of the proposed DiffDesign.

\textbf{Text Encoding Stage} 
In the text-to-image generation process of interior design, the encoding of text features plays a vital role as it converts natural language descriptions into numerical representations that computers can understand and process~\cite{qiao2020seed}. These encoded features will be used to guide the image generation model to generate images that match the descriptions. To obtain high-quality text representations, DiffDesign uses the CLIP text encoder~\cite{hafner2021clip,shen2021much} to extract features from text descriptions. CLIP is an important component for aligning image and text representations and can capture the correlation between cross-modal features. Since the open-source CLIP model cannot meet the requirements of professional terminology recognition in interior design generation, we use pre-trained CLIP for initialization and introduce a patch-based screening mechanism \cite{wang2023amsa} to locate keywords. Note that although CLIP may not fully capture specialized interior design terminology, it is used for its robust capability in aligning general visual and textual features. In our framework, CLIP is initialized in a frozen state and subsequently refined using a patch-based attention mechanism, which allows for more precise handling of interior design-specific concepts and terms during the generation process.
Specifically, we first collect a series of design proposals from interior design portfolios\footnote{We confirm that all datasets used in this study were obtained in full compliance with their respective Terms and Conditions. Specifically, we have only used datasets released for academic or non-commercial research purposes, and all usage adheres to the corresponding licenses and access policies.} based on the open-source projects, including the benchmark datasets with practical interior design cases, i.e., Matterport3D \cite{Chang_2017_Matterport3D}, HM3D \cite{Ramakrishnan_2021_HM3D}, and Zillow Indoor Dataset\cite{Cruz_2021_ZInD}, and interior design solutions designed by our team\footnote{This will be released in \href{https://github.com/Awesome-AutoDesign/DiffDesign}{https://github.com/Awesome-AutoDesign/DiffDesign} after being accepted}. We manually review text descriptions containing interior design-related terms. Next, we use a two-layer MLP trained on the pre-processed data and dynamically search for key regions based on the accumulated gradient heat map after dividing it into different patches. Briefly, for a new set of text descriptions, the trained network is able to calculate the corresponding weight matrix, where interior design terms are assigned higher weights.

\textbf{Text-to-image Generation Stage}
Once we successfully extract the features of the text, these features become the key to the entire generation process. They are cleverly integrated into a latent diffusion mechanism, which makes the generation process smoother and more efficient by simulating the natural distribution of data. In order to be able to generate images that are highly consistent with the input text description, we further subdivide this stage into three closely connected modules, each of which has specific tasks and responsibilities. The first is the Generative Appearance Control module, which is responsible for capturing and implementing the appearance features mentioned in the text description, such as color, shape, and basic composition elements. Through this module, the model learns how to transform the abstract description in the text into concrete visual effects in the image. Next is the Design Detail Control module, which focuses on the details in the text description, such as style, size range, and more complex visual elements. This module makes the generated image not only consistent with the text at the macro level, but also able to show rich details and depth at the micro level. Finally, there is the Optimization module, which fine-tunes and improves the outputs of the previous two modules to ensure that the generated image is as close to the text description as possible in terms of visuals, while also reaching high standards in terms of artistry and creativity. Through this module, the model is able to continuously learn and adapt to generate more detailed and realistic images. The whole process is a process of continuous iteration and optimization, with each module working on the basis of the previous module, ultimately achieving high-quality conversion from text to image. 

\subsection{Generated Appearance Control}
\label{sec:4.2}
To generate reliable interior design renderings, we decompose the task into explicitly disentangled control of generated appearance and design specifications. In this subsection, we introduce the generated appearance control process.

Specifically, we derive the semantic appearance context from the text encoding stage. This process involves extracting relevant features that capture the visual essence of the reference images. Once these features are obtained, we lock the feature extractor to ensure that these reference characteristics remain unchanged during the subsequent steps. Next, we activate the attention module, which provides layer-by-layer contextual guidance. This module integrates the derived appearance context into the generation process, ensuring that the synthesized images faithfully reflect the visual attributes of the reference examples while aligning with the provided textual descriptions. For this process, we integrate the appearance attributes of the reference text $c_{\text{text}}$ into the UNet backbone by activating the attention module to provide layer-by-layer contextual guidance.  To avoid the harmful effects of inaccurate text descriptions, we set $c_{\text{text}}$ with the assigned weight from the text encoding stage as the sole source of appearance. Formally, we denote the generative model as $f_{\theta}$ with parameters $\theta$, and the attention score is calculated as:
\begin{equation}\label{eq:attention}
    \text{Att}(Q, K, V) = \text{Softmax}\left(\frac{QK^T}{\sqrt{d_k}}\right)V,
\end{equation}
where $Q$, $K$, and $V$ are the query, key, and value features projected from the spatial features $h(z_t)$ using corresponding projection matrices, e.g., $Q=W_{q}h(z_t)$, $K=W_{k}h(z_t)$, and $V=W_{v}h(z_t)$.

To better guide the denoising process of Diffusion, $f_{\theta}$ is further copied into $f_{\theta}^{'}$ of trainable text-image pairs (data pairs used to train patch filtering mechanisms in text encoding), where $h^{'}(\cdot)$ is used as an intermediate representation in the UNet architecture. The design of $h$ ensures that the attention calculation is layer-by-layer and cross-query:
\begin{equation}
    \left\{\begin{matrix}
    Q^{\oplus } = W_q \cdot (h(z_t) \oplus h^{'} (z_t^{ref})),\\
    K^{\oplus } = W_k \cdot (h(z_t) \oplus h^{'} (z_t^{ref})),\\
    V^{\oplus } = W_v \cdot (h(z_t) \oplus h^{'} (z_t^{ref})),
    \end{matrix}\right.
\end{equation}
where $\oplus$ represents the direct sum of vector spaces. This step ensures that the model can cross-query relevant local content and textures from $\mathcal{D}^{ref}$ in addition to its own spatial features. Note that when more reference images are available (e.g. in a multi-view capture setting), this control module can be easily extended by concatenating multiple appearance contexts.

\subsection{Design Specification Control}
\label{sec:4.3}
In this subsection, we introduce the design specification control process. Specifically, we incorporate a design text description attention mechanism into the locked UNet decoder. The purpose of this module is to combine interior design text features with the image generation process, ensuring that the generated image not only retains the appearance features of the reference image but also accurately reflects the detailed terms and design specifications described in the text, such as the furniture size and ceiling wiring. It involves using design text features as queries, applying attention mechanisms to weigh the features within the decoder, and fusing these weighted features with the original features, thereby achieving precise control over the image generation.

Formally, given a design text description $c_{\text{text}}$, we encode it into text features via the text encoding stage mentioned in Subsection \ref{sec:4.1}. Let $f_{\text{decoder}}$ denote the locked UNet decoder, with the feature representation at the $l$-th layer denoted as $h^l(z_t)$. We use the design text features $c_{\text{design}}$ which is pre-processed by the patch-based selection scheme as the query, and the current layer's feature representation $h^l(z_t)$ as the keys and values, to compute the attention weights and apply them to the feature representation via the same function as Eq.\ref{eq:attention}. The projection matrices $W_q$, $W_k$, and $W_v$ are used to transform the respective features into the appropriate query, key, and value spaces. The attention mechanism calculates the weighted feature representation as follows:
\begin{equation}
    h_{\text{att}}^l(z_t) = \text{Attention}(W_q c_{\text{design}}, W_k h^l(z_t), W_v h^l(z_t))
\end{equation}
Next, we fuse the original feature representation $h^l(z_t)$ with the weighted feature representation $h_{\text{att}}^l(z_t)$. Assuming we use addition for fusion, the formula is:
\begin{equation}
h^l_{\text{fused}}(z_t) = h^l(z_t) + h_{\text{att}}^l(z_t)
\end{equation}
Finally, the fused feature representation $h^l_{\text{fused}}(z_t)$ is passed to the next layer, continuing the forward propagation in the UNet decoder:
\begin{equation}
h^{l+1}(z_t) = F_{\text{decoder}}^{l+1}(h^l_{\text{fused}}(z_t))
\end{equation}
This process ensures that the attention mechanism integrates the design text features at each layer, guiding the image generation in accordance with the specified design details. 

By integrating the design text description attention mechanism into the locked UNet decoder (design specification control module), we can precisely control the image generation process, ensuring that the generated image retains the reference image's appearance features and strictly follows the design text's professional terms and specifications.

\paragraph{More Details of Patch-based Selection}
The patch-based attention mechanism in our framework is an adaptation designed specifically for interior design tasks. While existing prompt-processing methods such as BLIP, Flamingo, and InstructPix2Pix process text descriptions in a linear manner, our approach divides the description into distinct patches that correspond to specific design elements. This enables our model to focus on more granular aspects of the design, such as layout, furniture type, and material specification, thereby enhancing the controllability and specificity of the generated outputs.

Specifically, as shown in Figure \ref{fig:1}, we define patches based on the spatial segmentation of the input text descriptions. The input text is parsed and divided into distinct semantic regions, where each patch corresponds to a region in the image that aligns with a particular design element or term in the prompt. For example, terms such as ``furniture'', ``walls'', or ``lighting'' are mapped to specific areas in the generated image that correspond to the respective elements of the interior design.
To identify the most important regions in the generated image, we utilize a two-layer MLP, denoted as $\text{MLP}_\text{patch}$. This network is trained to identify the significance of each patch by evaluating the gradient-based activations corresponding to each patch's design attribute. The gradients are computed with respect to the loss function during the backpropagation process, which helps determine the importance of different regions in relation to the overall design.
The output of the MLP consists of activation values for each patch, which are used to generate heatmaps. Each heatmap is a spatially distributed tensor, where high activation values correspond to regions of the image that need more focus. These heatmaps are then normalized to ensure that the total attention across all patches sums to one. We then apply the heatmaps to create a weight matrix, denoted as $W_\text{patch}$, which is used during the feature extraction process.
This weight matrix $W_\text{patch}$ is applied to the model’s internal feature maps at each layer to guide its attention to the regions identified as significant. The resulting adjusted feature maps are then passed through subsequent layers of the model, allowing it to focus more on the critical aspects of the design during the generation process. This strategy enables the model to generate more accurate and contextually relevant designs based on the given text description.

\subsection{Overall Optimization}
\label{sec:4.4}
To achieve precise control over the image generation process, we integrate the generated appearance control and design specification control as meta-priors into the diffusion model's optimization process. Below are the specific algorithmic formulas and detailed descriptions of the optimization process.

As mentioned in Subsection \ref{sec:3.2}, the generation objective of the Diffusion model is to minimize the reconstruction error in Eq.\ref{eq:loss}. In DiffDesign, the optimization objective further takes into account the control of generated appearance and design specification, where the model performs optimization by combining text conditions, reference image features, and design text features in the optimization of the diffusion model. Then, the overall optimization objective is defined as:
\begin{equation}
    \mathcal{L} = \mathbb{E}_{z_0, c_{\text{text}}, c_{\text{ref}}, c_{\text{design}}, t, \epsilon \sim \mathcal{N}(0,1)} \left[ \left\| \epsilon - \epsilon_\theta \left(z_t, c_{\text{text}}, c_{\text{ref}}, c_{\text{design}}, t\right) \right\|_2^2 \right].
\end{equation}
Here, $\epsilon_\theta$ is a parameterized U-Net architecture that takes encoded image latent $z_t$, text condition feature $c_{\text{text}}$, reference image feature $c_{\text{ref}}$ (the features obtained by $h^{'}(z_{t}^{ref})$), and design text feature $c_{\text{design}}$. Finally, we conduct controllable diffusion for efficient interior design generation.
Note that the fine-tuning process in DiffDesign does not fundamentally differ from other diffusion models in terms of adjusting the parameters based on the loss functions. What sets DiffDesign apart is the addition of the proposed module, e.g., control module, which enables it to handle more complex design specifications without requiring a complete overhaul of the fine-tuning framework.

\section{DesignHelper Dataset}
Interior design, a field blending creativity and functionality, has attracted increasing research attention due to advances in computer vision and generative models. However, the lack of specialized datasets for interior design generation and the limited diversity in existing datasets regarding space types, design styles, and solution complexity has hindered the development of automated interior design systems.

To address this gap and support the DiffDesign model, we present DesignHelper, a dataset designed to provide rich and diverse data for training interior design generation models. It includes over 15 space types and 15 design styles, covering more than 400 pairs of design solutions. This dataset aims to drive progress in the field of automated interior design.

The data for DesignHelper is sourced from public interior design platforms and contributions from professional designers, along with publicly available works from practitioners and students. By simulating various space requirements and styles, new design solutions were generated. This approach ensures diversity and representativeness in the dataset. To enhance quality, several interior design professionals reviewed and validated the designs to ensure practical feasibility.

\begin{figure*}[!h]
    \centering
    \includegraphics[width=\textwidth]{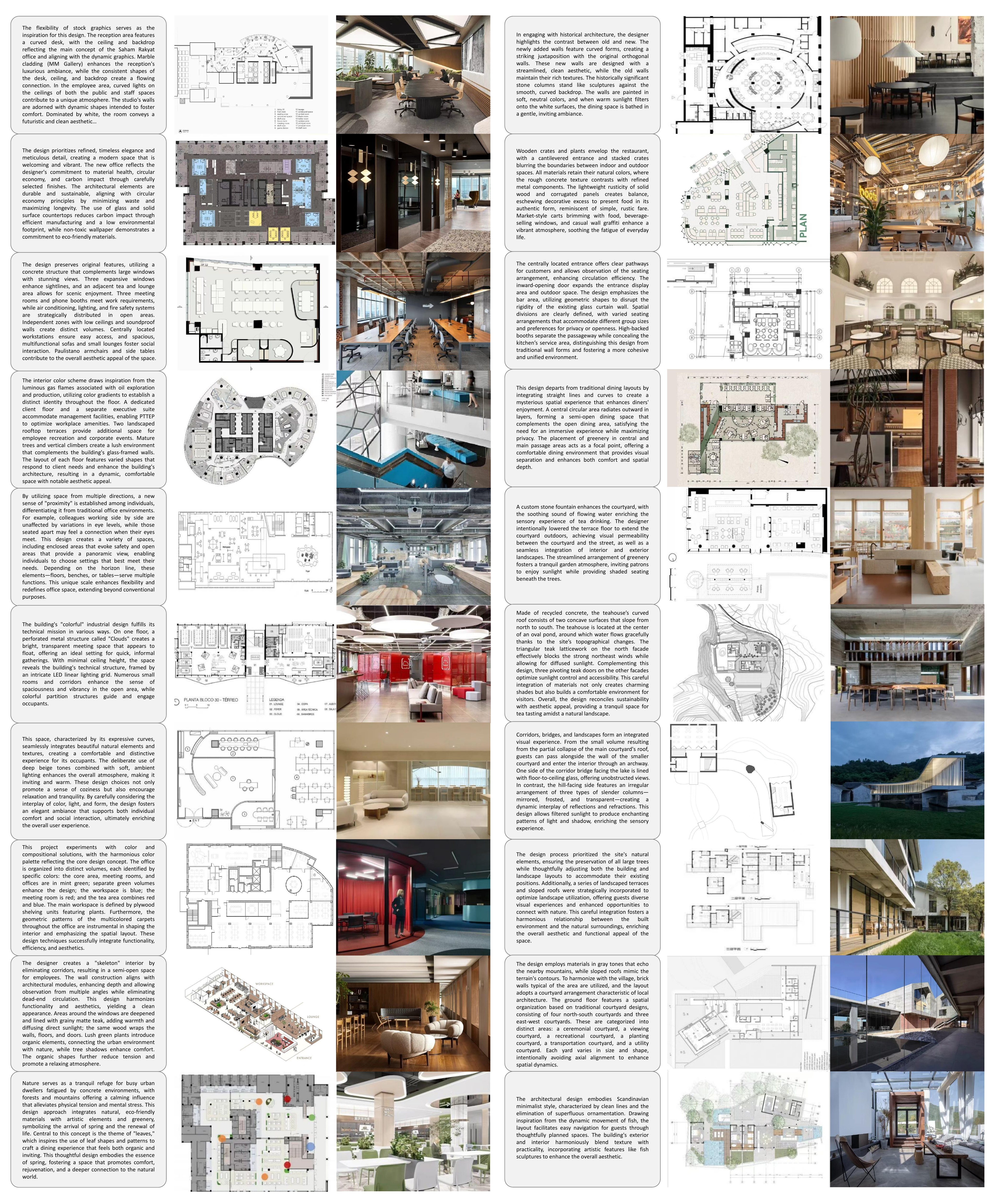}
    \caption{\textbf{DesignHelper dataset.} Here, we provide four interior design sample cases for public spaces, including offices, restaurants, teahouses, and hotels.}
    \label{fig:designhelper_1}
\end{figure*}

\begin{figure*}[!h]
    \centering
    \includegraphics[width=\textwidth]{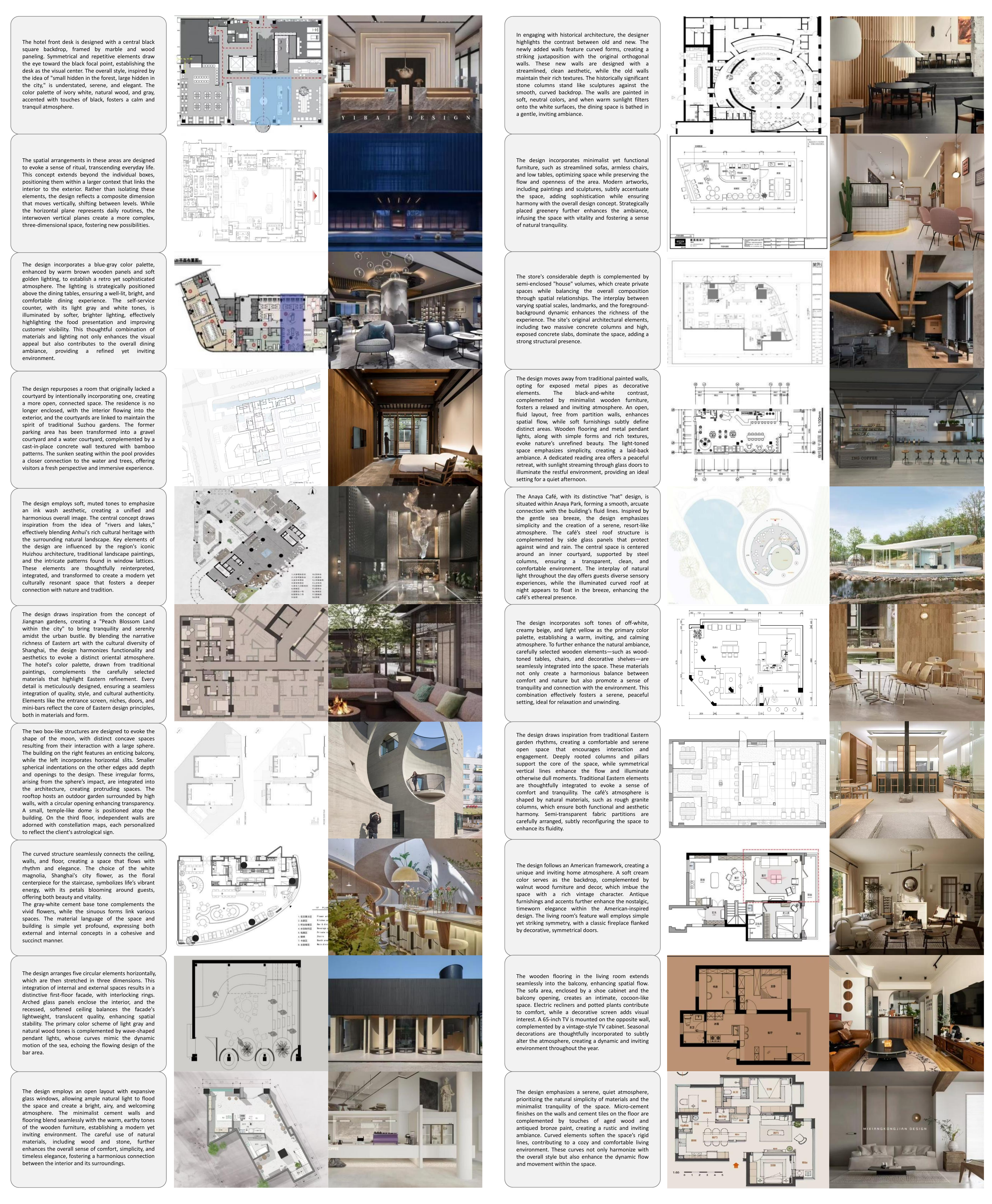}
    \caption{\textbf{DesignHelper dataset.} Here, we provide four interior design sample cases for home and leisure spaces, including cafe, gallery, living room, and book bar.}
    \label{fig:designhelper_2}
\end{figure*}

DesignHelper includes over 15 space types and 15 design styles, covering more than 400 design pairs. The space types span residential, commercial, office, and entertainment contexts, each reflecting distinct functional requirements and design challenges. These include living rooms, kitchens, bedrooms, bathrooms, studies, dining rooms, offices, cafes, hotel rooms, shops, and exhibition halls. The dataset features 15 design styles, such as modern, minimalist, Nordic, industrial, vintage, rustic, Japanese, Mediterranean, luxurious, classic, classical, futuristic, eco-friendly, eclectic, and country, reflecting current interior design trends and diverse usage scenarios. Each design pair includes a description, floor plan, and rendering. Images are in PNG format with a 1024×1024 resolution for efficient model training. Design descriptions are in JSON format, detailing space type, style, color scheme, furniture layout, and decoration style, which serve as prompts for generating floor plans and renderings. Examples of DesignHelper's spaces are shown in Fig~\ref{fig:designhelper_1} and Fig~\ref{fig:designhelper_2}. The dataset is published at \href{https://github.com/Awesome-AutoDesign/DiffDesign}{https://github.com/Awesome-AutoDesign/DiffDesign}, but permission is required to download the full content.

\section{Experiments}
\label{sec:5}
In this section, we first introduce the experimental setup in Subsection \ref{sec:5.1}. Then, we present the results of qualitative evaluation, quantitative evaluation, and computational efficiency evaluation in Subsections \ref{sec:5.2} to \ref{sec:5.4}, demonstrating the superiority of DiffDesign. Next, we visualize some results of DiffDesign and provide the human preference evaluation in Subsection \ref{sec:5.5}. Finally, we conduct ablation studies to further explore how DiffDesign works well in Subsection \ref{sec:5.6}. 

\subsection{Experimental Setup}
\label{sec:5.1}
In this subsection, we introduce implementation details, evaluation protocols, and baselines of our experiments in turn.

\subsubsection{\textbf{Implementation Details}}
\label{sec:5.1.1}
The fine-tuning diffusion model is implemented using PyTorch. We set the training process to 0.5 million iterations, with each session lasting 32 hours. The pre-processing method automatically resized input images to a resolution of $512 × 512$ pixels. 
We utilize the Adam optimizer \cite{zhang2018improved} for training our model. We set momentum and weight decay to 0.9 and $10^{-4}$, respectively. The initial learning rate for all experiments is set at $2\times 10^{-5}$ with the flexibility for linear scaling as needed, while the batch size is set to 8. All reported metrics (FID, IS, and CLIP Sim) are computed as averages over three independent runs, each with different random seeds. This ensures that the results reflect the general performance of the model rather than being dependent on any single run.

\subsubsection{\textbf{Evaluation Protocols}}
\label{sec:5.1.2}
The evaluation protocol combines both machine and human evaluations, i.e., Subsection \ref{sec:5.2}-\ref{sec:5.4} based on machine evaluation, Subsection \ref{sec:5.5} based on human evaluation, and Subsection \ref{sec:5.6} is based on both, to provide a comprehensive understanding of the model’s performance. Machine evaluation metrics include CLIP performance with image-to-text and text-to-image retrieval, CLIP Similarity (CLIP Sim) for measuring semantic alignment between generated images and text descriptions, Inception Score (IS) for assessing image quality and diversity, and Fréchet Inception Distance (FID) for evaluating the distance between distributions of generated and real images \cite{hafner2021clip,gan2023idesigner}. Human evaluation involves subjective assessments by a group of evaluators who rate the images based on visual appeal, relevance to the prompts, and overall aesthetic quality. This dual approach ensures a well-rounded evaluation by integrating objective computational metrics with human perceptual judgments.

\subsubsection{\textbf{Baselines}}
\label{sec:5.1.3}
We consider five strong baselines for generate effect evaluation: Midjourney~\cite{borji2022generated}, DALL-E 3~\cite{betker2023improving}, Stable Diffusion~\cite{rombach2022high}, iDesigner~\cite{gan2023idesigner}, and SD-XL~\cite{podell2023sdxl}. Midjourney is an advanced AI art generator that focuses on delivering visually stunning and imaginative creations.
DALLE 3 is renowned for its innovative text-to-image capabilities, generating high-quality images from textual descriptions. It serves as a benchmark for cutting-edge generative models. 
Stable Diffusion is another classic Diffusion model that can generate high-quality, high-resolution images based on a given text description.
iDesigner is a framework specific for generating interior design solutions based on the Diffusion model.
SD-XL, on the other hand, is a variant of the Stable Diffusion model known for its extended capabilities in handling complex image synthesis tasks. 
It is worth noting that, considering the capabilities of different components of DiffDesign, we also analyzed its computational efficiency and cross-modal adaptation separately, and selected different baseline methods for comparison, which are described in the corresponding subsections.
Through a comparative analysis of DiffDesign with recognized models, our goal is to highlight the efficacy and innovative aspects of our methodology, with a focus on its performance in generating bilingual images and its responsiveness to textual cues.

\subsection{Qualitative Evaluation}
\label{sec:5.2}
To verify the model's effectiveness, we use fixed text descriptions for different indoor spaces to generate corresponding interior design schemes. We select five areas: a five-star hotel lobby, a modern café, a study room, a living room, and a clothing store showroom. Each area has a specific decorating style and detailed requirements. For example, the hotel lobby features a Neo-Chinese style with a reception desk, viewing area, and lounge area; the café has a modern luxury style with a bar counter and booths; the showroom is a children's themed clothing store with bright, lively colors; the living room follows a cream-colored style with a TV, coffee table, and sofa in a simple and bright design; and the study room has a tech-filled style with a desk, bookshelves, and many books. Note that the descriptions for each space include both rough and detailed versions. The rough text description contains only key information such as furniture styles, while the detailed description includes design details such as furniture dimensions and ceiling structures. For example, a large L-shaped sectional sofa with a chaise lounge on one end, upholstered in soft neutral-toned fabric, with a seat height of 18 inches and a seat depth of 22 inches.
Next, we adopt the same experimental setting and conduct experiments on different baselines and our DiffDesign for performance comparison.

\begin{figure*}[!h]
    \centering
    \includegraphics[width=\textwidth]{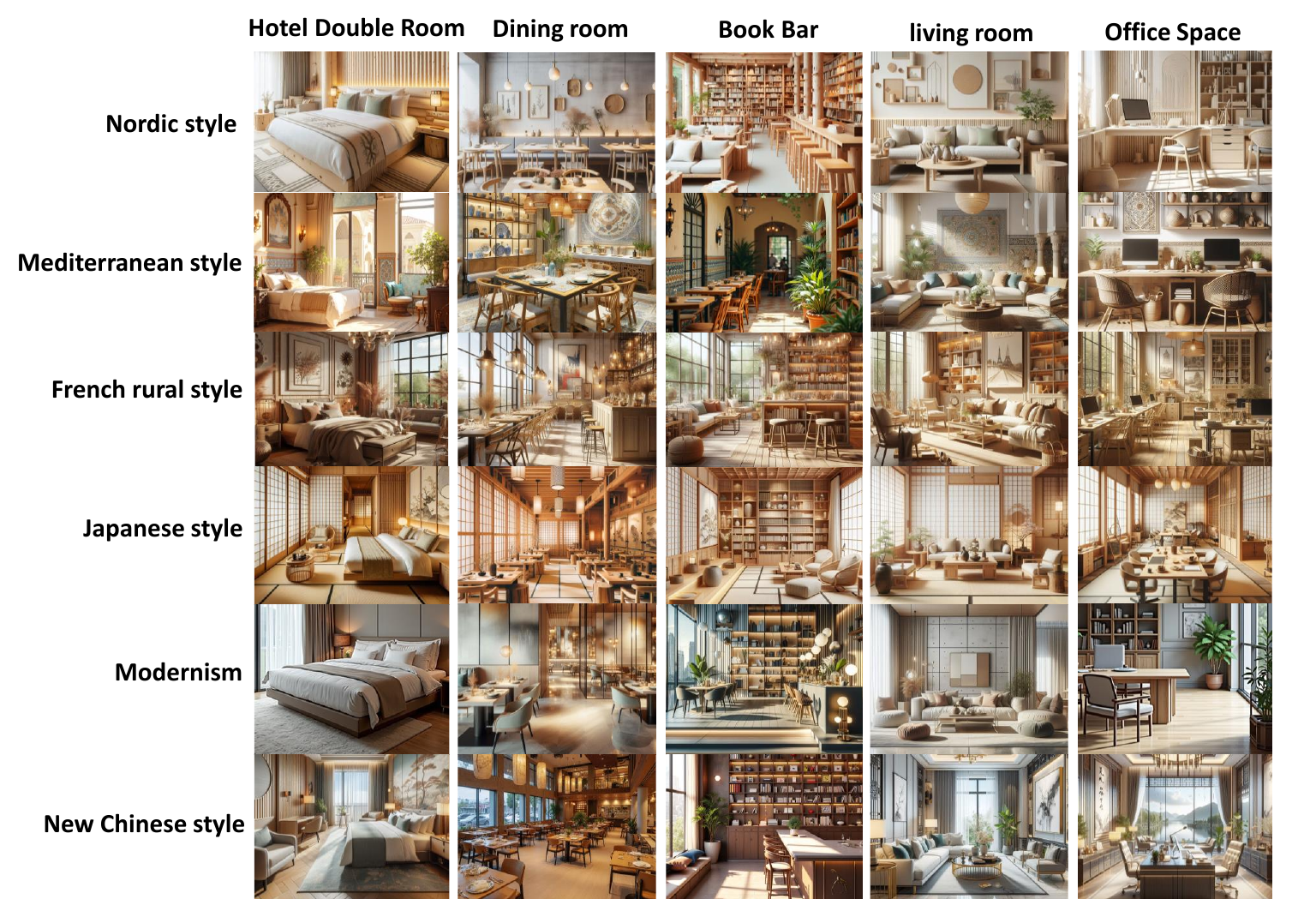}
    \caption{\textbf{Interior design images generated by our diffusion model for different decoration styles and space functions.}}
    \label{fig:ex_1_1}
\end{figure*}

\begin{figure*}[!h]
    \centering
    \includegraphics[width=\textwidth]{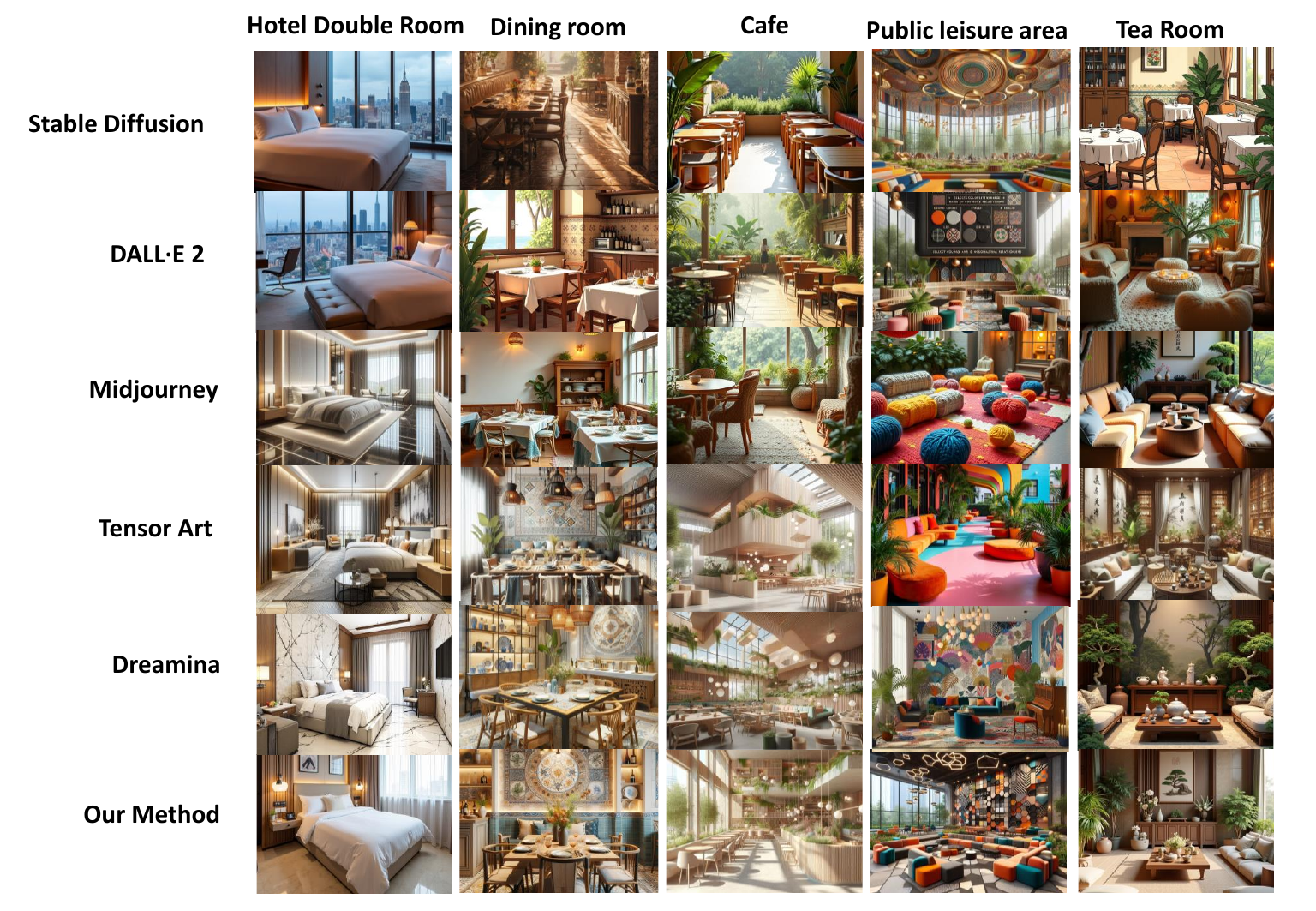}
    \caption{\textbf{Interior design images generated by different diffusion models for different space functions.}}
    \label{fig:ex_1_2}
\end{figure*}

The results in Fig~\ref{fig:ex_1_1} show that our method achieves high-fidelity and 3D consistent synthesis of new views at a resolution of 512 × 512. Meanwhile, although our method only involves training on aligned image text, it shows excellent generalization ability to new indoor areas, styles, and design sizes.
Furthermore, Fig~\ref{fig:ex_1_2} shows the comparison results of some state-of-the-art new-generation frameworks in generating interior design solutions from text. The results show that our method is far superior to previous studies in terms of both generation quality and satisfaction of design requirements. For example, The ceiling perspectives rendered by Tensor Art are inaccurate, creating a sense of visual oppression or dissonance. Such errors can make spaces appear smaller and adversely affect the overall design. Similarly, the proportions between furniture, decorations, and spaces generated by Stable Diffusion are often unrealistic, leading to designs that look unnatural and do not meet practical usage requirements. Furthermore, DALL·E 2 fails to adequately depict critical details such as outlets, switches, and vents. In contrast, the solutions generated by DiffDesign can reasonably arrange the fluidity and functional layout of the space while accurately expressing the texture and color of the materials.

\subsection{Quantitative Evaluation}
\label{sec:5.3}
To evaluate the effectiveness of DiffDesign, the quantitative experiments are divided into two parts: a comparative experiment for CLIP performance and an assessment of DiffDesign's generative performance. Specifically, for CLIP, we select the Flickr~\cite{young2014image}, MSCOCO~\cite{lin2014microsoft}, and DesignHelper as benchmark datasets and perform cross-language image-text retrieval tasks. For DiffDesign, we conduct comparative experiments on the English dataset following~\cite{gan2023idesigner} and evaluate using CLIP Similarity (CLIP Sim), Inception Score (IS), and Fréchet Inception Distance (FID) metrics.

Table \ref{tab:ex2_1_1} and Table \ref{tab:ex2_1_2} show the retrieval performance of the pre-trained CLIP model on the three datasets, respectively. The results indicate that our CLIP model achieves the best performance on almost all the datasets. On the Flickr and MSCOCO datasets, our CLIP achieves comparable results to AltCLIP and iDesign-CLIP models, which demonstrate a better understanding than the original CLIP. On the DesignHelper dataset, our CLIP achieves an optimal accuracy of 85.9\%. Table \ref{tab:ex2_2} provides a comprehensive overview of DiffDesign and other baselines' performance on the English dataset. The results also prove the outstanding effectiveness and robustness of DiffDesign, achieving the best results.

\begin{table*}[!ht]
\centering
\caption{Zero-shot image-text retrieval results on Flickr and MSCOCO datasets. The best results are marked in \textbf{bold}.}
\label{tab:ex2_1_1}
\resizebox{\textwidth}{!}{%
\begin{tabular}{@{}lcccccccccccc@{}}
    \toprule 
    & \multicolumn{6}{c}{Flickr} & \multicolumn{6}{c}{MSCOCO} \\
    & \multicolumn{3}{c}{Image $\rightarrow$ Text} & \multicolumn{3}{c}{Text $\rightarrow$ Image} & \multicolumn{3}{c}{Image $\rightarrow$ Text} & \multicolumn{3}{c}{Text $\rightarrow$ Image} \\
    \cmidrule(lr){2-4} \cmidrule(lr){5-7} \cmidrule(lr){8-10} \cmidrule(lr){11-13}
    Model & R@1 & R@5 & R@10 & R@1 & R@5 & R@10 & R@1 & R@5 & R@10 & R@1 & R@5 & R@10 \\
    \midrule
    CLIP  & 85.1 & 97.3 & 99.2 &  65.0 & 87.1 & 92.2 &  56.4 & 79.5 & 86.5 &  36.5 & 61.1 & 71.1 \\
    SD-XL & 84.2 & 96.1 & 97.4 & 64.3 & 85.9 & 91.7 & 55.4 & 78.2 & 85.8 & 34.5 & 60.4 & 69.6 \\
    CLIPOOD & 85.7 & 98.2 & 99.8 & 67.4 & 88.3 & 93.5 & 57.9 & 81.7 & 88.1 & 36.9 & 63.4 & 73.8 \\
    AltCLIP & 86.0 & 98.0 & 99.1 &  72.5 & 91.6 & 95.4 &  58.6 & 80.6 & 87.8 &  42.9 & 68.0 & 77.4 \\
    iDesignerCLIP   & 88.4 & 98.8 & \textbf{99.9} & 75.7 & 93.8 & 96.9 & \textbf{61.2} & 84.8 & 90.3 & 49.2 & 70.3 & 79.6  \\
    DiffDesign & \textbf{90.2} & \textbf{99.4} & 99.7 & \textbf{78.7} & \textbf{94.2} & \textbf{97.6} & 60.3 & \textbf{85.9} & \textbf{92.4} & \textbf{49.6} & \textbf{74.1} & \textbf{80.5} \\
    \bottomrule
\end{tabular}
}
\end{table*}

\begin{table*}[!ht]
\centering
\caption{Zero-shot image-text retrieval results on DesignHelper datasets. The best results are marked in \textbf{bold}.}
\label{tab:ex2_1_2}
\resizebox{\textwidth}{!}{%
\begin{tabular}{@{}lcccccccccccc@{}}
    \toprule 
    & \multicolumn{6}{c}{DesignHelper-Graphic Design} & \multicolumn{6}{c}{DesignHelper-Renderings} \\
    & \multicolumn{3}{c}{Image $\rightarrow$ Text} & \multicolumn{3}{c}{Text $\rightarrow$ Image} & \multicolumn{3}{c}{Image $\rightarrow$ Text} & \multicolumn{3}{c}{Text $\rightarrow$ Image} \\
    \cmidrule(lr){2-4} \cmidrule(lr){5-7} \cmidrule(lr){8-10} \cmidrule(lr){11-13}
    Model & R@1 & R@5 & R@10 & R@1 & R@5 & R@10 & R@1 & R@5 & R@10 & R@1 & R@5 & R@10 \\
    \midrule
    CLIP   & 67.5 & 74.9 & 77.0 & 40.1 & 52.8 & 55.4 & 78.1 & 85.1 & 88.0 & 44.2 & 64.8 & 71.5 \\
    SD-XL & 68.7 & 76.9 & 79.2 & 42.6 & 54.3 & 56.8 & 79.1 & 86.7 & 90.4 & 46.1 & 67.3 & 72.5 \\
    CLIPOOD & 64.8 & 71.6 & 74.2 & 38.7 & 49.1 & 52.8 & 75.0 & 82.1 & 84.7 & 40.9 & 61.5 & 68.3 \\
    AltCLIP  & 69.4 & 77.5 & 81.4 & 45.2 & 54.6 & 59.1 & 80.2 & 87.9 & 91.3 & 48.2 & 67.6 & 74.4 \\
    iDesignerCLIP  & 68.5 & 75.3 & 79.4 & 43.2 & 54.8 & 57.9 & 79.3 & 86.1 & 90.7 & 46.3 & 65.9 & 73.8 \\
    DiffDesign & \textbf{74.9} & \textbf{82.4} & \textbf{84.6} & \textbf{49.3} & \textbf{59.6} & \textbf{61.7} & \textbf{85.9} & \textbf{91.6} & \textbf{94.8} & \textbf{50.2} & \textbf{69.8} & \textbf{77.4} \\
    \bottomrule
\end{tabular}
}
\end{table*}

\begin{table}[!ht]
\centering
\caption{Comparison of different models based on CLIP Sim and IS and FID across English dataset. The best results are marked in \textbf{bold}.}
\label{tab:ex2_2}
\resizebox{0.7\textwidth}{!}{%
\begin{tabular}{lccc}
\toprule
Model & CLIP Sim($\uparrow$)  & IS($\uparrow$) & FID($\downarrow$) \\
\cmidrule(r){1-4}
\multicolumn{4}{c}{English Dataset} \\
\midrule
Test Set & 0.205 & 4.838 & 0 \\
SD-XL & 0.112 & 3.450 & 95.867 \\
DALL-E 3 & 0.118  & 3.832 & 89.906 \\
iDesigner\_1024 & 0.135  & 4.562 & 79.340 \\
iDesigner\_2048 & 0.137  & 4.559 & 79.262 \\
iDesigner\_RLCF & 0.145  & 4.690 & 76.832 \\
DiffDesign & \textbf{0.174 $\pm$ 0.002} & \textbf{4.720 $\pm$ 0.104} & \textbf{75.118 $\pm$ 0.010}\\
\bottomrule
\end{tabular}}
\end{table}

\subsection{Computational Efficiency Evaluation}
\label{sec:5.4}
To analyze the computational efficiency of DiffDesign, we construct comparative experiments between DiffDesign and baselines to evaluate their trade-off between performance and efficiency. For this purpose, we develop an edge computing setup utilizing the NVIDIA Jetson AGX Xavier as the central unit. This platform is equipped with a 512-core NVIDIA Volta GPU accompanied by 64 Tensor cores, a pair of NVIDIA deep learning accelerators, a duo of vision accelerators, and an octa-core NVIDIA Carmel Arm CPU. We record the training time of each model under different accuracy effects. It is worth noting that the frameworks we compare here are all diffusion and corresponding variants, and experiments are conducted on the built DesignHelper to evaluate the effect of our work. The selected baselines include Stable Diffusion\cite{ho2020denoising}, Q-Diffusion\cite{bremner2022qdiffusion}, Score SDE\cite{song2021score}, LDM\cite{rombach2022high}, iDDPM\cite{nichol2021improved}, ADM\cite{dhariwal2021diffusion}, VDM\cite{kingma2021variational}, iDesigner\cite{xu2023idesigner}, DiffDesign \cite{zhang2023diffdesign}.

\begin{figure}[!h]
    \centering
    \includegraphics[width=0.6\linewidth]{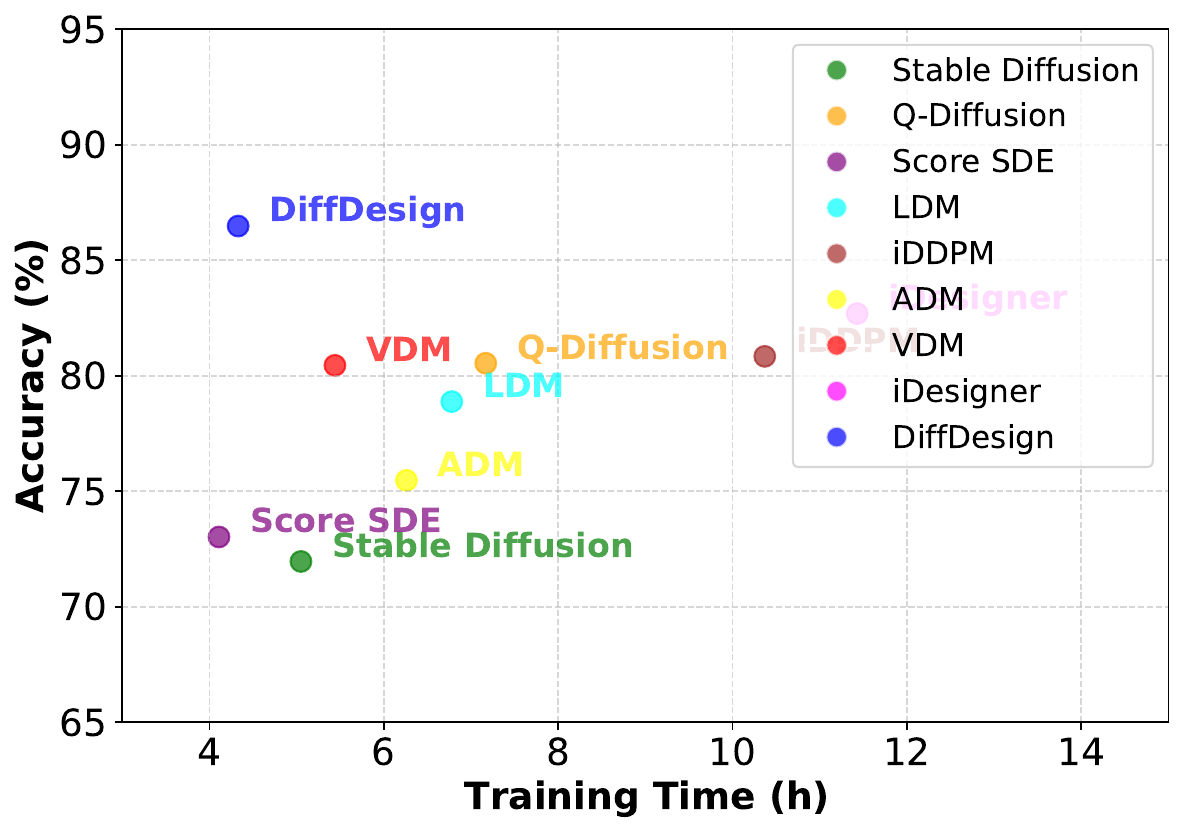}
    \caption{\textbf{Computational Efficiency Evaluation.} The results shows the training time of different models, reflecting the trade-off performance.}
    \label{fig:ex3}
\end{figure}

Fig~\ref{fig:ex3} illustrates the results of computational efficiency, i.e., training time. The center of each circle in the figure represents the average result for the respective models, and the circle's area represents a 90\% confidence interval. Combining the results in Table \ref{tab:ex2_1_1}, Table \ref{tab:ex2_1_2}, and Table \ref{tab:ex2_2}, we observe that on the same platform, DiffDesign achieves fast training while maintaining great performance. This result highlights the superiority of DiffDesign in terms of computational efficiency.

\subsection{Human Preference Evaluation}
\label{sec:5.5}
In addition to automated metrics, we conducted a manual preference evaluation to directly assess the perceptual quality of the generated images. We recruited 20 interior design researchers and professionals for the experiment, using six methods—DiffDesign, Midjourney, DALL-E 3, Stable Diffusion, iDesigner, and SD-XL—to create spatial renderings and floor plans based on 20 interior design prompts. Each image was rated on a five-point quality scale, and we recorded the average preference score for each method across participants. To mitigate potential bias, participants were not informed of the source model for each image. The human preference evaluation was conducted using the following criteria: (i) Design Appeal: How visually appealing is the generated design? (ii) Task Relevance: How well does the design align with the input prompt specifications? (iii) Usability: How practical is the design for real-world implementation? These criteria were chosen because they reflect both subjective visual quality and objective design utility, making them effective in evaluating the real-world applicability of generated designs.

\begin{figure}[!h]
    \centering
    \includegraphics[width=0.6\linewidth]{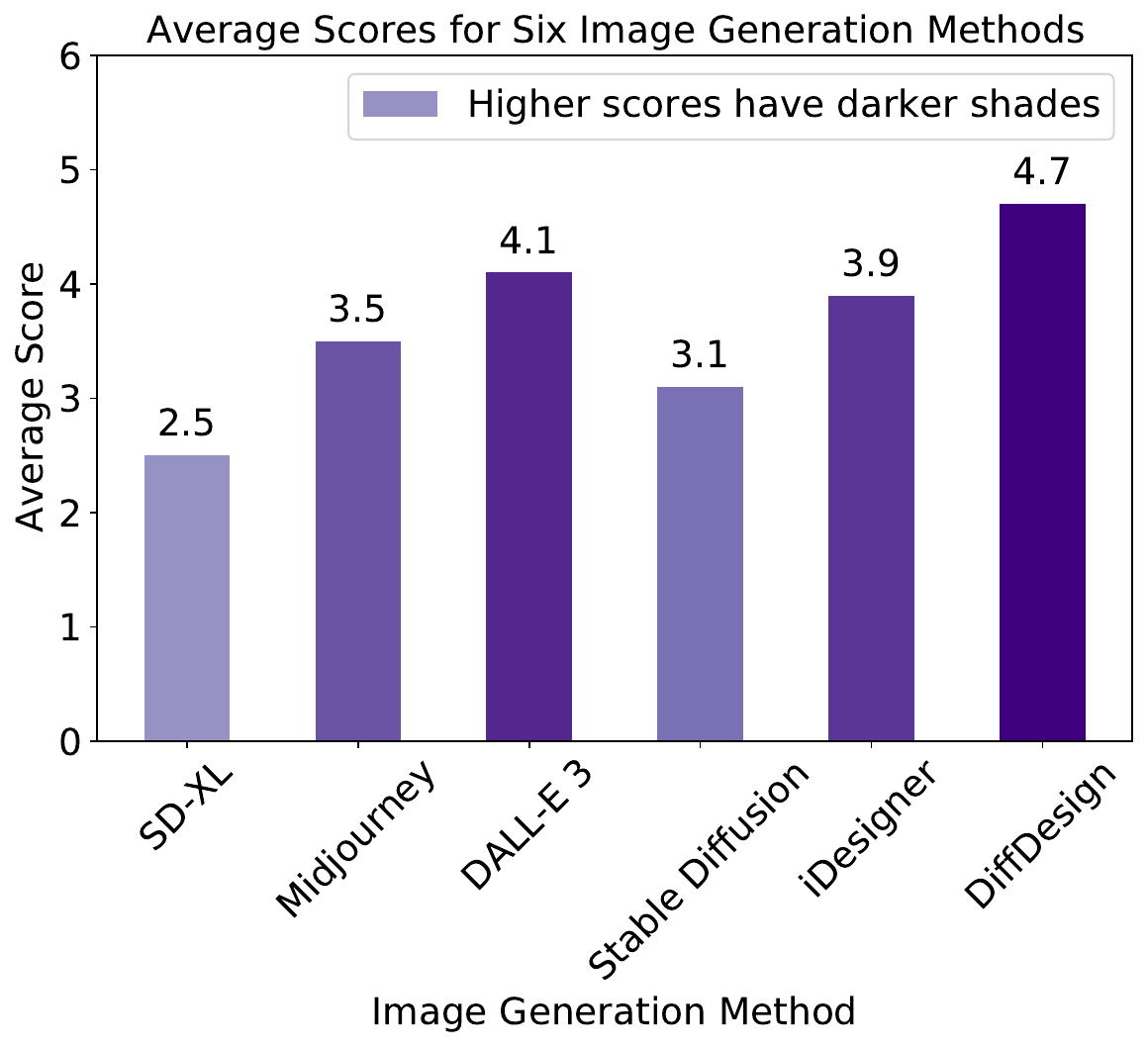}
    \caption{\textbf{Average scores of human preference for six interior design solutions generation models.}}
    \label{fig:ex_human_1}
\end{figure}

\begin{figure*}[!h]
    \centering
    \includegraphics[width=\linewidth]{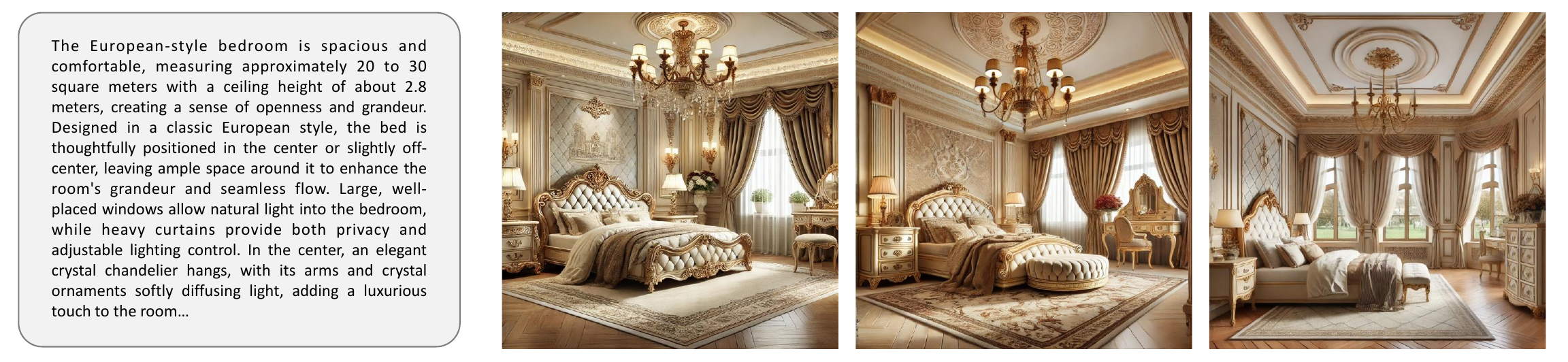}
    \caption{\textbf{Renderings of different models generated based on the same prompt.} It requires the generation of a bedroom space in a European classical style, while constraining information such as space size, ceiling, lighting, wall material, etc. We can observe that only the sample generated by DiffDesign meets this requirement (\textbf{the first one}), while the other two have problems such as perspective confusion and asymmetric windows.}
    \label{fig:ex_human_2}
\end{figure*}

The results, depicted in Fig~\ref{fig:ex_human_1}, clearly indicate a preference for DiffDesign, with the model achieving a win rate of over 62\% against other baselines. This significant margin not only reinforces the quantitative findings but also illustrates the qualitative leap in generation. The generated images are frequently cited as more coherent, aesthetically pleasing, and true to the textual descriptions provided. In Fig~\ref{fig:ex_human_2}, we present images generated by the same prompt using different models.
This subjective evaluation underscores the effectiveness of DiffDesign in preserving the semantic essence of the original prompts while manifesting images that resonate more strongly with human judges. The feedback from human evaluators provides invaluable insights that go beyond numerical scores, highlighting the nuanced improvements that DiffDesign brings to the realm of text-to-image synthesis.

\subsection{Ablation Study}
\label{sec:5.6}
We conduct an ablation study to explore the impact of different modules in DiffDesign, i.e., the generated appearance control module $M_{ga}$, the design specification control module $M_{ds}$, and the interior-design-specific CLIP $M_{clip}$. We evaluate their influence on the DiffDesign performance by replacing or redesigning these modules. For instance, $M_{ga}$ and $M_{ds}$ are disabled, and $M_{clip}$ is substituted with pre-trained CLIP without the text training mentioned in Subsection \ref{sec:4.1}.

\begin{figure*}[!h]
    \centering
    \includegraphics[width=\linewidth]{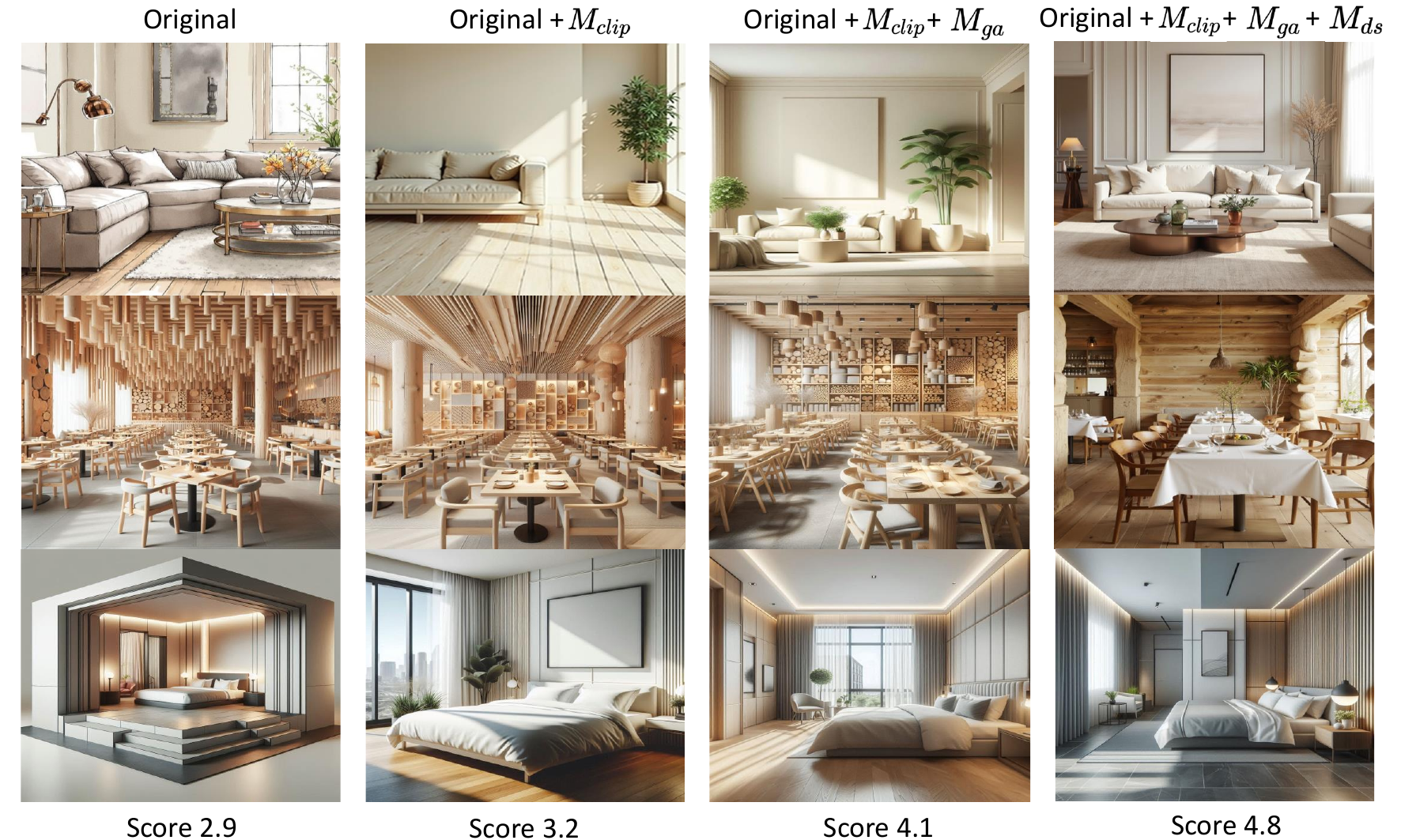}
    \caption{\textbf{Ablation study of the three modules of the proposed method, i.e., $M_{clip}$, $M_{ga}$, and $M_{ds}$.} We select three sets of text descriptions representing different spaces—living room, dining room, and bedroom—and include ratings from interior design professionals for the generated samples in each set.}
    \label{fig:ex_abla_1}
\end{figure*}

The results are shown in Fig~\ref{fig:ex_abla_1}. 
Volunteers are recruited to score generated samples, with scores shown below each sample. The results demonstrate that the three modules substantially improve model performance. Specifically, \(M_{ga}\) and \(M_{clip}\) enhance the style of the generated interior design renderings, especially in cartoon vs. photorealistic effects, furniture layout, and ceiling perspective. Similarly, the addition of \(M_{ds}\) allows the renderings to better capture lighting and shadow variations aligned with user preferences, meeting the actual needs of users. 
Therefore, our design demonstrates foresight, while highlighting our concern for the issue of spatial and temporal heterogeneity in fine-grained emotion recognition.

\section{Discussion}

In this study, we introduce DiffDesign, a controllable diffusion model enhanced with a meta prior, designed to address key challenges in interior design generation. This section provides a brief critical analysis of our findings, comparing them with existing research in the field and highlighting the contributions and innovations made in this study.

Table \ref{tab:comparison_discussion} presents a summary of the most important results from our study, comparing them with findings from previous studies in generative models under the context of interior design. Unlike previous models, which struggle with high fidelity and design-specific control, DiffDesign provides more refined control over both appearance and design specifications. The disentangled cross-attention mechanism for appearance and design specifications allows for precise adjustments, ensuring that generated designs meet the required standards for real-world applications. Meanwhile, DiffDesign introduces DesignHelper, a specialized dataset designed specifically for interior design. Unlike datasets used in previous works like DALL-E 3 and Stable Diffusion, which rely on general image datasets, DesignHelper offers tailored design solutions across various spatial types and design styles. This contributes to the model’s ability to generate contextually accurate interior design renderings, ensuring both visual quality and functional design compliance. Thus, DiffDesign achieves not only high diversity but also ensures that generated outputs match the detailed specifications provided in the prompts, enhancing its practical utility for interior design professionals.

\begin{table}[h!]
\centering
\caption{A brief critical analysis of DiffDesign.}
\label{tab:comparison_discussion}
\resizebox{\textwidth}{!}{%
\begin{tabular}{l|l|l|l|l|l}
\toprule
\textbf{Model} & \textbf{Quality} & \textbf{Controllability} & \textbf{Diversity} & \textbf{Efficiency} & \textbf{Dataset} \\ 
\midrule
\textbf{Midjourney} & Moderate & Low & High & High & General image datasets \\ 
\textbf{DALL-E 3} & High & Moderate & Moderate & Moderate & General image datasets \\
\textbf{Stable Diffusion} & Moderate & Low & Moderate & Moderate & General image datasets \\ 
\textbf{iDesigner} & High & Moderate & Moderate & Low & Interior design-specific dataset \\ 
\textbf{SD-XL} & High & Low & High & Moderate & General image datasets \\ 
\textbf{DiffDesign} & High & High & High & Efficient & DesignHelper \\ 
\bottomrule
\end{tabular}
}
\end{table}

While DiffDesign demonstrates outstanding performance, there are areas where improvements can be made. For instance, the model’s performance on highly complex or highly specific design requirements, such as the integration of sustainability features or advanced ergonomic needs, could be further explored. Moreover, enhancing the diversity of the DesignHelper dataset by including more varied design types and user-generated content could improve the model’s robustness and adaptability to different design contexts.

\section{Conclusion}
\label{sec:6}
In this study, we address the inefficiencies and limitations of traditional interior design generation methods, which often fail to meet practical needs and provide controllable quality. To overcome these challenges, we introduce DiffDesign, a controllable diffusion model enhanced with a meta prior, designed for efficient interior design generation. By leveraging a pre-trained 2D diffusion model as the rendering backbone and introducing disentangled cross-attention control for design attributes such as appearance, pose, and size, DiffDesign enables high-quality, customizable outputs. The inclusion of an optimal transfer-based alignment module ensures that generated designs maintain view consistency. Through extensive experiments on various benchmark datasets, we demonstrate that DiffDesign not only outperforms existing methods in terms of prediction accuracy but also enhances computational efficiency. Additionally, the introduction of the DesignHelper dataset, which contains over 400 design solutions across 15 spatial types and 15 design styles, significantly contributes to the fine-tuning and validation of the model. The implications of this work extend beyond interior design, offering a promising approach to generative models in fields that require a high degree of design control and customization. Note that while DiffDesign demonstrates strong performance in interior design generation, its effectiveness for interior design is currently tied to the DesignHelper dataset and the quality of input prompts. Future work will focus on further refining DiffDesign’s ability to integrate real-time feedback and support even more complex design scenarios, paving the way for broader adoption in practical applications.

\section*{Acknowledgments}
This research received no external funding.

\nolinenumbers

%
%
%




\bibliography{references}

\end{document}